\documentclass{article}
\usepackage{amsfonts,amsmath,amsthm,amssymb}

\date{}

\usepackage{algorithm}
\usepackage{algorithmic}
\usepackage{subfigure}
\usepackage{epsfig}
\usepackage{graphicx}
\usepackage{url}
\usepackage{multirow}
\usepackage{amssymb}
\usepackage{mathrsfs}
\usepackage{epstopdf}
\usepackage{multicol}
\usepackage{amsmath}
\usepackage{mathtools}

\usepackage{multicol}
\usepackage{wrapfig,lipsum,booktabs}
\usepackage{bm}
\usepackage{boxedminipage}
\usepackage{xcolor}

\usepackage{multirow}

\usepackage{pifont}

\allowdisplaybreaks[4]

\newtheorem{theorem}{\textbf{Theorem}}
\newtheorem{assumption}{\textbf{Assumption}}

\newtheorem{corollary}{\textbf{Corollary}}
\newtheorem{remark}{\textbf{Remark}}

\numberwithin{equation}{section}

\theoremstyle{plain}

\theoremstyle{definition}

\title{Achieving Linear Speedup in Decentralized Stochastic Compositional Minimax Optimization}

\author{
	Hongchang Gao \thanks{Temple University, {\tt hongchang.gao@temple.edu}} 
	}
\begin{document}
\maketitle

%
\begin{abstract}
	The stochastic compositional minimax problem has attracted a surge of attention in recent years since it covers many emerging machine learning models. Meanwhile, due to the emergence of distributed data, optimizing this kind of problem under the decentralized setting becomes badly needed. However, the compositional structure in the loss function brings unique challenges to designing efficient decentralized optimization algorithms. In particular, our study shows that the standard gossip communication strategy cannot achieve linear speedup for decentralized compositional minimax problems due to the large consensus error about the inner-level function. To address this issue, we developed a novel decentralized stochastic compositional gradient descent ascent with momentum algorithm to reduce the consensus error in the inner-level function. As such, our theoretical results demonstrates that it is able to achieve linear speedup with respect to the number of workers. We believe this novel algorithmic design could benefit the development of decentralized compositional optimization.   Finally, we applied our methods to the imbalanced classification problem. The extensive experimental results provide evidence for the effectiveness of our algorithm. 
\end{abstract}

\section{Introduction}
In this paper, we consider the decentralized stochastic compositional minimax optimization problem:
\begin{equation} \label{loss}
	\begin{aligned}
		& \min_{\mathbf{x}\in\mathbb{R}^{d_{1}}} \max_{\mathbf{y}\in \mathbb{R}^{d_{2}}} \frac{1}{K}\sum_{k=1}^{K} f_k(g_k(\mathbf{x}), \mathbf{y})  \ .
	\end{aligned}
\end{equation}
Specifically, there are $K$ workers and each worker  communicates with its neighbors in a decentralized manner. Moreover,  each worker $k$ ($k\in \{1,2,\cdots, K\}$) has its own loss function  $f_k(g_k(\mathbf{x}), \mathbf{y})$, which is a compositional function with respect to $\mathbf{x}$. More specifically, the inner-level function is defined as $g_k(\mathbf{x})=\mathbb{E}_{\xi_k}[g_k(\mathbf{x}; \xi_k)]: \mathbb{R}^{d_{1}} \rightarrow \mathbb{R}^{d_0}$, where $\xi_k$ represents the data distribution on the $k$-th worker for the inner-level function. The outer-level function is defined as $f_k(g_k(\mathbf{x}), \mathbf{y})=\mathbb{E}_{\zeta_k}[f_k(g_k(\mathbf{x}), \mathbf{y}; \zeta_k) ]: \mathbb{R}^{d_0}\times \mathbb{R}^{d_{2}}\rightarrow \mathbb{R}$, where $\zeta_k$ denotes the data distribution on the $k$-th worker for the outer-level function.

The stochastic compositional minimax optimization problem in Eq.~(\ref{loss}) covers many machine learning problems, such as  distributionally robust stochastic compositional optimization problems \cite{gao2021convergence},  stochastic compositional AUROC maximization problems \cite{yuan2021compositional}.  Thus, it has been actively studied and applied to numerous machine learning models in the past few years. However, the compositional structure in the loss function incurs significant challenges to optimize it. Specifically, the stochastic gradient $\nabla_{\mathbf{x}} f_k(g_k(\mathbf{x}; \xi_k), \mathbf{y}; \zeta_k)$  regarding $\mathbf{x}$ is not an unbiased estimation for the full gradient  due to the  compositional structure in the objective function.  This biased estimator causes new challenges for both \textit{the update on individual workers} and \textit{the communication across workers}.

To address the challenge caused by the compositional structure in the loss function, a lot of efforts have been made under the single-machine setting (i.e., the \textit{update on individual workers}) in the past few years. For instance, \cite{wang2017stochastic} developed the stochastic compositional gradient descent method for  stochastic compositional \textit{minimization} problems. After that, numerous  methods have been developed to improve the convergence rate \cite{zhang2019composite,yuan2019stochastic,zhang2019stochastic,chen2020solving,yuan2020stochastic}. However, all of them only concentrate on the compositional \textit{minimization} problem. It is unclear whether they can converge for  compositional \textit{minimax} problems.  Recently, \cite{gao2021convergence}  developed a stochastic compositional gradient descent ascent method for optimizing Eq.~(\ref{loss}) under the single-machine setting. More recently, \cite{yuan2021compositional} proposed a primal-dual stochastic compositional adaptive method, which also concentrates on the update on individual workers. Thus, these optimization methods cannot be used to optimize  Eq.~(\ref{loss}) since they are not able to address the challenges regarding the \textit{communication across workers} caused by the compositional structure. 

As for the decentralized setting, numerous decentralized optimization methods have been developed to optimize large-scale machine learning models in recent years.  For instance, \cite{lian2017can} developed stochastic gradient descent for nonconvex optimization problems for the \textit{minimization} problem. Afterwards, a wide variety of methods \cite{lu2019gnsd,pu2020distributed,koloskova2019decentralized,tang2019deepsqueeze,sun2020improving,lin2021quasi,yuan2021decentlam,koloskova2021improved}  have been proposed to improve the computation complexity and communication complexity, and some works \cite{xian2021faster} developed algorithms for the minimax optimization problem.  However, all these methods only concentrate on the problem without  the compositional structure. Thus, they fail to  address the unique challenges regarding  \textit{the update on individual workers} and \textit{the communication across workers} for  compositional minimax optimization problems.  Recently, a couple of parallel compositional optimization algorithms  have been proposed under either the decentralized \cite{gao2021fast,zhao2022distributed,yang2022decentralized} or centralized setting \cite{tarzanagh2022fednest,gao2022convergence,wang2021memory}. However, all of them only concentrate on the compositional minimization problem. Additionally, they fail to achieve the linear speedup except \cite{gao2021fast,yang2022decentralized}, where \cite{gao2021fast} requires a large batch size while \cite{yang2022decentralized} requires an unrealistic operation, i.e., communicating the high dimensional $\nabla g_k(\mathbf{x}; \xi_k)\in \mathbb{R}^{d_{0}\times d_{1}}$.   Therefore, \textit{how to address the challenges about the update on individual workers and the communication across workers to achieve the \textbf{linear speedup} for decentralized \textbf{compositional minimax} problems} is still an open problem.

\vspace{-5pt}
\subsection{Our Contributions}
\vspace{-5pt}
The contributions of this paper lie in the development of novel 
decentralized  stochastic compositional gradient descent ascent methods and the establishment of the linear speedup convergence rate, which are summarized below. 
\begin{itemize}
	\item We first developed a  decentralized stochastic compositional gradient descent ascent with momentum  algorithm based on the gossip communication strategy (D-SCGDAM-GP).  We found D-SCGDAM-GP can only achieve the $O(\frac{1}{\epsilon^4})$ iteration/communication complexity to find the $\epsilon$-accuracy solution, failing to achieve the linear speedup regarding the number of workers, i.e.,  $O(\frac{1}{K\epsilon^4})$. The reason is that the large consensus error of the inner-level function prohibits D-SCGDAM-GP  from achieving that.

	\item Based on our findings in the first algorithm, we developed a novel decentralized stochastic compositional gradient descent ascent with momentum  algorithm based on the gradient-tracking communication strategy (D-SCGDAM-GT). In particular, D-SCGDAM-GT communicates the inner-level function value to control the consensus error so that it is able to theoretically achieve linear speedup regarding the number of workers.  To the best of our knowledge, this is the first algorithm  communicating the inner-level function. We believe this novel algorithmic design can benefit the development of  other decentralized compositional optimization problems. 
	
	\item The communication of the inner-level function causes new challenges for convergence analysis. We successfully addressed these challenges and established the convergence rate of our D-SCGDAM-GT algorithm. In particular, it  can achieve the $O(\frac{1}{K\epsilon^4})$ iteration/communication complexity and  $O(\frac{1}{K\epsilon^4})$ sample complexity for  under the nonconvex-strongly-concave problem.  We believe our theoretical analysis strategy regarding the design of communicating the inner-level function can also benefit the  development of  decentralized compositional optimization.

	\item Finally, we applied our proposed methods to the decentralized compositional AUROC maximization problem for imbalanced data classification. The extensive experimental results on multiple benchmark datasets confirm the superior performance of our proposed methods. 
\end{itemize}

\section{Related Work}
\subsection{Stochastic Compositional Minimization and Minimax Problems}
\vspace{-5pt}
The stochastic compositional minimization problem has been extensively studied in the past few years due to its widespread application in machine learning models, e.g., sparse additive models \cite{wang2017stochastic}, model-agnostic meta-learning \cite{finn2017model}. To alleviate the bias of  stochastic gradients caused by the compositional structure in the loss function, \cite{wang2017stochastic} developed a stochastic compositional gradient descent (SCGD) method, which utilizes a moving average strategy to reduce the estimation variance of the inner-level function. As such, SCGD can achieve the  sample complexity $O(\frac{1}{\epsilon^8})$ for nonconvex problems. Afterwards, a series of methods have been proposed to improve its sample complexity. For instance, \cite{yuan2019stochastic} utilized the SPIDER gradient estimator \cite{fang2018spider} so that the sample complexity is improved to $O(\frac{1}{\epsilon^{3}})$ . However, this method requires to periodically use a large batch size of samples to compute the gradient, which is not practical for large-scale data.  Recently, \cite{ghadimi2020single} developed a momentum stochastic compositional   gradient descent (SCGDM) method, whose sample complexity is $O(\frac{1}{\epsilon^4})$ with a small batch size.  Thus, this method is efficient for large-scale data and it has been applied to various applications, such as AUPRC maximization problem \cite{qi2021stochastic}. However, all these methods can only optimize the stochastic compositional \textit{minimization} problem, rather than the \textit{minimax compositional}   problem in Eq.~(\ref{loss}).   

As for the \textit{compositional minimax} problem, a typical application is Area Under the Curve (AUROC) maximization.  It is an effective method for the imbalanced classification problem. Its goal is to directly optimize the AUROC score, instead of the cross-entropy loss function. Recently, under the\textit{ single-machine setting}, to enable the stochastic training of AUROC maximization  problems,  \cite{ying2016stochastic} reformulates AUROC maximization as a minimax optimization problem under the \textit{single-machine setting} as follows:
\vspace{-10pt}
\begin{equation} \label{auc}
	\begin{aligned}
		& \min_{\bm{\theta}, \hat{\theta}_1, \hat{\theta}_2}\max_{\tilde{\theta}} \frac{1}{n}\sum_{i=1}^{n} \mathcal{L}_i(\bm{\theta}, \hat{\theta}_1, \hat{\theta}_2, \tilde{\theta}) \ , 
	\end{aligned}
	\vspace{-5pt}
\end{equation}
where  
\begin{equation}
	\begin{aligned}
		& \mathcal{L}_i(\bm{\theta}, \hat{\theta}_1, \hat{\theta}_2, \tilde{\theta})=(1-p)(f(\bm{\theta} ; \mathbf{a}_{i})-\hat{\theta}_1)^{2} \mathbb{I}_{[b_{i}=1]} +p(f(\bm{\theta} ; \mathbf{a}_{i})-\hat{\theta}_2)^{2} \mathbb{I}_{[b_{i}=-1]}  -p(1-p) \tilde{\theta}^{2}\\
		& \quad \quad \quad \quad \quad \quad \quad  +2(1+\tilde{\theta} )(p f(\bm{\theta}; \mathbf{a}_{i}) \mathbb{I}_{[b_{i}=-1]}-(1-p) f(\bm{\theta}; \mathbf{a}_{i}) \mathbb{I}_{[b_{i}=1]}) \ .
	\end{aligned}
\end{equation}
Here, $\bm{\theta}\in \mathbb{R}^d$ denotes the model parameter of the classifier $f(\bm{\theta} ; \mathbf{a}_{i})$, $\hat{\theta}_1\in \mathbb{R}$ and $\hat{\theta}_2 \in \mathbb{R}$ are two additional parameters in the minimization subproblem, $\tilde{\theta} \in \mathbb{R}$ is the parameter for the maximization subproblem, $(\mathbf{a}_{i}, b_{i})$ is the input sample, and $p$ is the ratio of positive samples. When the classifier is a nonconvex function, such as deep neural network, Eq.~(\ref{auc}) is a nonconvex-strongly-concave minimax problem, which can be efficiently optimized by stochastic gradient descent ascent. 
However, its empirical performance is not satisfied. Recently, to address this problem, \cite{yuan2021compositional} developed a compositional AUROC maximization method. In particular, it is to optimize a compositional loss function, where the outer-level function is an AUROC loss function and the inner-level function is induced by a cross-entropy loss function. As such, it becomes a nonconvex-strongly-concave compositional minmax  optimization problem:
\begin{equation} \label{compositional_auc}
	\begin{aligned}
		& \min_{\bm{\theta}, \hat{\theta}_1, \hat{\theta}_2}\max_{\tilde{\theta}} \frac{1}{n}\sum_{i=1}^{n} \mathcal{L}_i(\bm{\theta} -\frac{\rho}{n}\sum_{j=1}^{n}\nabla \mathcal{L}_{j}^{cr}(\bm{\theta}), \hat{\theta}_1, \hat{\theta}_2, \tilde{\theta} )\ , 
	\end{aligned}
\end{equation}
where $\mathcal{L}_j^{cr}(\bm{\theta})$ denotes the cross-entropy loss function for the $j$-th sample, $\rho>0$ is the step size. Then, the inner-level function is $g(\bar{\bm{\theta}}) = \bar{\bm{\theta}} - \rho\Delta(\bar{\bm{\theta}})$ where $\bar{\bm{\theta}} = [\bm{\theta}^T, \hat{\theta}_1, \hat{\theta}_2]^T$ and $\Delta(\bar{\bm{\theta}}) = [\frac{1}{n}\sum_{j=1}^{n}(\nabla \mathcal{L}_{j}^{cr}(\bm{\theta}))^T, 0, 0]^T$, and the outer-level function is $f(g(\bar{\bm{\theta}} ), \tilde{\theta})=\frac{1}{n}\sum_{i=1}^{n}\mathcal{L}_i(g(\bar{\bm{\theta}} ), \tilde{\theta})$.

Recently,  \cite{gao2021convergence} studied the stochastic \textit{compositional} minimax problem under the single-machine setting, i.e.,  $K=1$ in Eq.~(\ref{loss}),  and developed the stochastic compositional gradient descent ascent (SCGDA) method. Even though SCGDA can achieve the $O(\frac{1}{\epsilon^4})$ sample complexity when the loss function is  nonconvex-strongly-concave, its batch size should be as large as $O(\frac{1}{\epsilon^2})$, which is not practical in real-world applications. More recently, \cite{yuan2021compositional} developed the primal-dual stochastic compositional adaptive (PDSCA) method to solve Eq.~(\ref{compositional_auc}). It can achieve the same sample complexity as SCGDA with a constant batch size. However, both SCGDA and PDSCA only concentrate on the single-machine setting. Thus, it is unclear whether they will converge under the decentralized setting.

\vspace{-5pt}
\subsection{Decentralized Stochastic Compositional Optimization Problems} 
The decentralized optimization method has been extensively studied in recent years. Typically, there are two communication strategies: the gossip strategy and gradient-tracking strategy. The gossip method only communicates the model parameter, while the gradient tracking method communicates both model parameters and gradients. Based on these two communication strategies, numerous decentralized optimization methods have been explored. For instance, \cite{lian2017can} established the convergence rate of the gossip-based decentralized stochastic gradient descent (DSGD) method for nonconvex problems, while \cite{lu2019gnsd} studied the convergence rate of  gradient-tracking-based DSGD. In addition, some efforts have been made to improve the sample complexity and communication complexity by compressing the communicated variables \cite{tang2019deepsqueeze,koloskova2019decentralized} or reducing the gradient variance \cite{sun2020improving,zhang2021gt}. However, all of these methods only concentrate on the \textit{minimization} problem. Even though some efforts \cite{xian2021faster,tsaknakis2020decentralized} have been made for the decentralized minimax optimization methods, they only concentrate on the \textit{non-compositional} problem.  
Thus, they cannot be used to optimize Eq.~(\ref{loss}) and their theoretical analyses are not applicable to our methods. 
As for the compositional optimization problem, \cite{tarzanagh2022fednest,gao2022convergence,wang2021memory} developed stochastic compositional gradient descent methods for minimization problems under the centralized federated learning setting. All of them fail to achieve linear speedup. Recently, \cite{zhang2023federated} developed a centralized federated compositional minimax optimization algorithm. However, due to the periodic global communication, the consensus error is reset to zero periodically so that it is much easier to achieve linear speedup than the decentralized setting, where the consensus error is always non-zero. Under the decentralized setting, \cite{gao2021fast,zhao2022distributed,yang2022decentralized} developed  a couple of algorithms that only work for the compositional minimization problem. Even though \cite{gao2021fast,yang2022decentralized} can achieve linear speedup, they require unrealistic operations. Specifically, \cite{gao2021fast} requires that the batch size should be as large as $O(\frac{1}{\epsilon^2})$, while  \cite{yang2022decentralized} requires to communicate a very high dimensional matrix $\nabla g_k(\mathbf{x}; \xi_k)\in \mathbb{R}^{d_{0}\times d_{1}}$. Taking Eq.~(\ref{compositional_auc}) as an example, if we use ResNet50 as the classifier where  $d_{0}=d_{1}$ is as large as  25 million, it is impossible to communicate $d_{1}^2$ parameters in each iteration in a real-world system. 
In summary, all existing methods are not capable of optimizing Eq.~(\ref{loss}), and their theoretical analyses are not applicable to our settings. Thus, it is necessary to develop new methods with rigorous theoretical guarantees to optimize Eq.~(\ref{loss}).

\section{Problem Setup}
We make the following commonly used assumptions for the stochastic compositional optimization problem \cite{wang2017stochastic,ghadimi2020single,gao2021convergence}. 
\begin{assumption} \label{assumption_smooth}
	For any outer-level function $f_k(\cdot, \cdot)$,  it has $L_f$-Lipschitz continuous gradient, i.e.,  for $\forall (\mathbf{x}_1, \mathbf{y}_1), (\mathbf{x}_2, \mathbf{y}_2)\in \mathbb{R}^{d_1}\times \mathbb{R}^{d_2}$,   $ \|\nabla_{g} f_k(g_k(\mathbf{x}_1), \mathbf{y}_1)-\nabla_{g}  f_k(g_k(\mathbf{x}_2), \mathbf{y}_2)\|^2 \leq L_f^2(\|g_k(\mathbf{x}_1)-g_k(\mathbf{x}_2)\|^2  + \|\mathbf{y}_1- \mathbf{y}_2\|^2 )$ and $\|\nabla_{\mathbf{y}}  f_k(g_k(\mathbf{x}_1), \mathbf{y}_1)-\nabla_{\mathbf{y}}  f_k(g_k(\mathbf{x}_2), \mathbf{y}_2)\|^2 \leq  L_f^2(\|g_k(\mathbf{x}_1)-g_k(\mathbf{x}_2)\|^2  + \|\mathbf{y}_1- \mathbf{y}_2\|^2 ) $,
	where $L_f>0$.  For any inner-level function $g_k(\cdot)$,  it has $L_g$-Lipschitz continuous gradient, i.e.,  for $\forall \mathbf{x}_1, \mathbf{x}_2 \in \mathbb{R}^{d_1}$,  $ \|\nabla g_k(\mathbf{x}_1) - \nabla g_k(\mathbf{x}_2)\|^2 \leq L_g^2 \|\mathbf{x}_1-\mathbf{x}_2\|^2$,
	where $L_g>0$.

\end{assumption}

\begin{assumption} \label{assumption_bound_gradient}
	For any outer-level function $f_k(\cdot, \cdot)$ and inner-level function $g_k(\cdot)$, $\forall (\mathbf{x}, \mathbf{y}) \in \mathbb{R}^{d_1}\times\mathbb{R}^{d_2}$,   their stochastic gradients satisfy: $ \mathbb{E}[\|\nabla_{g} f_k(g_k(\mathbf{x}), \mathbf{y}; \zeta)\|^2] \leq C_f^2$, $\mathbb{E}[\|\nabla_{\mathbf{y}} f_k(g_k(\mathbf{x}), \mathbf{y}; \zeta)\|^2] \leq C_f^2$, and $\mathbb{E}[\|\nabla g_k(\mathbf{x}; \xi)\|^2]\leq C_g^2$, 
	where $C_f>0$ and $C_g>0$ are two constants.  
\end{assumption}

\begin{assumption} \label{assumption_bound_variance}
	For any outer-level function $f_k(\cdot, \cdot)$ and  $\forall (\mathbf{x}, \mathbf{y}) \in \mathbb{R}^{d_1}\times\mathbb{R}^{d_2}$, the variance of its stochastic gradient satisfies: $ \mathbb{E}[\|\nabla_{g} f_k(g_k(\mathbf{x}), \mathbf{y}; \zeta) - \nabla_{g} f_k(g_k(\mathbf{x}), \mathbf{y})\|^2] \leq \sigma_f^2$ and $\mathbb{E}[\|\nabla_{\mathbf{y}} f_k(g_k(\mathbf{x}), \mathbf{y}; \zeta) - \nabla_{\mathbf{y}} f_k(g_k(\mathbf{x}), \mathbf{y})\|^2] \leq \sigma_f^2$, 
	where $\sigma_f>0$.
	For any inner-level function $g_k(\cdot)$ and $\forall \mathbf{x} \in \mathbb{R}^{d_1}$,  the variance of its stochastic gradient and function value satisfy: $\mathbb{E}[\|\nabla g_k(\mathbf{x}; \xi) - \nabla g_k(\mathbf{x})\|^2] \leq \sigma_{g'}^2 $ and $\mathbb{E}[\| g_k(\mathbf{x}; \xi) -  g_k(\mathbf{x})\|^2] \leq \sigma_g^2$, 
	where $\sigma_g>0$ and $\sigma'_g>0$. 
\end{assumption}

\begin{assumption} \label{assumption_strong}
	For any function $f_k(g_k(\mathbf{x}), \mathbf{y})$, it is $\mu$-strongly-concave regarding $\mathbf{y}$
	where $\mu>0$. 
\end{assumption}

\begin{assumption} \label{assumption_graph}
	The adjacency matrix $W=[w_{ij}]\in \mathbb{R}^{K\times K}$ of the graph that composed by all workers is a symmetric and doubly stochastic matrix. Its second largest absolute eigenvalue $\lambda$ satisfies $\lambda<1$.  
\end{assumption}

In terms of the aforementioned assumptions, we denote the condition number as $\kappa=L_f/\mu$ and the spectral gap as $1-\lambda$. Moreover, throughout this paper,  we use $\mathbf{a}_{k, t}$ to denote the variable $\mathbf{a}$ on the $k$-th worker at the $t$-th iteration, and  denote    $\bar{\mathbf{a}}_t=\frac{1}{K}\sum_{k=1}^{K}\mathbf{a}_{k,t}$. We further
introduce the auxiliary function $\Phi_k(\mathbf{x}) = \max_{\mathbf{y}\in \mathbb{R}^{d_2}} f_k(g_k(\mathbf{x}), \mathbf{y})$  and $\mathbf{y}_k^*(\mathbf{x}) = \arg \max_{\mathbf{y}\in \mathbb{R}^{d_2}} f_k(g_k(\mathbf{x}), \mathbf{y})$, where  $\Phi_k(\mathbf{x})$ is $L_{\Phi}$-smooth with $L_{\Phi}=2C_g^2L_f^2/\mu+ C_fL_g$ \cite{gao2021convergence}.
Additionally, we represent the loss function as $F(\mathbf{x}, \mathbf{y})=\frac{1}{K}\sum_{k=1}^{K}F_k(\mathbf{x}, \mathbf{y})= \frac{1}{K}\sum_{k=1}^{K} f_k(g_k(\mathbf{x}), \mathbf{y})$. Then, $\min_{\mathbf{x}\in\mathbb{R}^{d_1}} \max_{\mathbf{y}\in \mathbb{R}^{d_2}} F(\mathbf{x}, \mathbf{y}) = \min_{\mathbf{x}\in\mathbb{R}^{d_1}} \Phi(\mathbf{x})$.  where $\Phi(\mathbf{x}) = \frac{1}{K}\sum_{k=1}^{K}\Phi_k(\mathbf{x})$.

\begin{algorithm}[h]
	\caption{D-SCGDAM-GP}
	\label{alg_dscgdamgp}
	\begin{algorithmic}[1]
		\REQUIRE $\mathbf{x}_0$, $\mathbf{y}_0$, $\eta\in (0, 1)$, $\gamma_x>0$, $\gamma_y>0$, $\beta_x>0$, $\beta_y>0$, $\alpha>0$,  $\alpha\eta \in (0, 1)$, $\beta_x\eta \in (0, 1)$, $\beta_y\eta \in (0, 1)$. \\

		\FOR{$t=0,\cdots, T-1$, each worker $k$} 
		\IF {$t==0$}
		\STATE 
		$\mathbf{x}_{k,0}=\mathbf{x}_0$,\   $\mathbf{y}_{k,0}=\mathbf{y}_0$, \ 
		$\mathbf{h}_{k, 0}= g_k(\mathbf{x}_{k, 0};  \xi_{k, 0})$, \\
		$\mathbf{u}_{k, 0}=\nabla g_k(\mathbf{x}_{k, 0};  \xi_{k, 0})^T\nabla_{g}  f_k(\mathbf{h}_{k,0}, \mathbf{y}_{k,0}; \zeta_{k,0})$, \ 
		$\mathbf{v}_{k,0}=\nabla_{y}  f_k(\mathbf{h}_{k,0}, \mathbf{y}_{k,0}; \zeta_{k,0})$,
		\ENDIF
		\STATE 
		$\tilde{\mathbf{x}}_{k, t+1} = \sum_{j: w_{kj}> 0}w_{kj}\mathbf{x}_{j,t} - \gamma_x \mathbf{u}_{k, t}$,  \ 
		$\mathbf{x}_{k, t+1}= \mathbf{x}_{ k,t} + \eta(\tilde{\mathbf{x}}_{k, t+1}-\mathbf{x}_{ k,t} )$,
		\STATE 
		$\tilde{\mathbf{y}}_{k, t+1}  = \sum_{j: w_{kj}> 0}w_{kj}\mathbf{y}_{j,t}+ \gamma_y \mathbf{v}_{k, t}$,  \ 
		$\mathbf{y}_{k, t+1}= \mathbf{y}_{ k,t} + \eta(\tilde{\mathbf{y}}_{k, t+1} -\mathbf{y}_{ k,t} )$,
		
		%

		\STATE 
		$\mathbf{h}_{k, t+1} = (1- \alpha\eta) \mathbf{h}_{k, t} +  \alpha\eta g_k(\mathbf{x}_{k,t+1};  \xi_{k, t+1})$, \\
		$\mathbf{u}_{k, t+1} = (1-\beta_x\eta)\mathbf{u}_{k, t} + \beta_x\eta \nabla g_k(\mathbf{x}_{k, t+1};  \xi_{k, t+1})^T\nabla_{g}  f_k(\mathbf{h}_{k,t+1}, \mathbf{y}_{k,t+1}; \zeta_{k,t+1})$,\\
		$\mathbf{v}_{k, t+1} = (1-\beta_y\eta)\mathbf{v}_{k, t} + \beta_y\eta\nabla_{y}  f_k(\mathbf{h}_{k,t+1}, \mathbf{y}_{k,t+1}; \zeta_{k,t+1})$,
		\ENDFOR
	\end{algorithmic}
\end{algorithm}

\section{Gossip-based Algorithm}

In this section, we developed a decentralized algorithm based on the gossip communication strategy, which employs the  stochastic compositional gradient descent with momentum algorithm to address \textit{the challenge regarding the update on individual workers caused by the compositional structure}. 

\subsection{Algorithmic Design}
In Algorithm~\ref{alg_dscgdamgp}, we developed the gossip-based decentralized  stochastic compositional gradient descent ascent with momentum (D-SCGDAM-GP) algorithm.  Generally speaking, there are $K$ workers, where each worker conducts local updates and then communicates the local model parameter with its neighboring workers. In detail, the model parameters $\mathbf{x}_{k, 0}$ and $\mathbf{y}_{k, 0}$ on all workers are intilized with the same value $\mathbf{x}_0$ and  $\mathbf{y}_{0}$, respectively.  Then, for the minimization subproblem, at the $t$-th iteration, the $k$-th worker computes the stochastic compositional gradient based on the estimation for the inner-level function $g_k(\mathbf{x})$ as below:
\begin{equation}
	\begin{aligned}
		& \mathbf{h}_{k, t} = (1- \alpha\eta) \mathbf{h}_{k, t-1} +  \alpha\eta g_k(\mathbf{x}_{k,t};  \xi_{k, t}) \ ,
	\end{aligned}
\end{equation}
where $\alpha$ and $\eta$ are two positive hyperparameters, $\xi_{k,t}$ denotes the selected samples on the $k$-th worker at the $t$-th iteration. Here, $\mathbf{h}_{k, t}$ can be viewed as the moving average estimation for the inner-level function $g_k(\mathbf{x})$ when $\alpha\eta<1$. This strategy is commonly used in stochastic compositional gradient descent method \cite{wang2017stochastic}. Based on $\mathbf{h}_{k, t}$, our method computes the momentum as below:
\begin{equation}
	\begin{aligned}
		& \mathbf{u}_{k, t} = (1-\beta_x\eta)\mathbf{u}_{k, t-1}+ \beta_x\eta \nabla g_k(\mathbf{x}_{k, t};  \xi_{k, t})^T\nabla_{g}  f_k(\mathbf{h}_{k,t}, \mathbf{y}_{k,t}; \zeta_{k,t}) \ ,
	\end{aligned}
\end{equation}
where $\beta_x$ and $\eta$ are two positive hyperparameters and $\beta_x\eta<1$,  $\mathbf{u}_{k, t}$ denotes the momentum for the stochastic compositional gradient $\nabla g_k(\mathbf{x}_{k, t};  \xi_{k, t})^T\nabla_{g}  f_k(\mathbf{h}_{k,t}, \mathbf{y}_{k,t}; \zeta_{k,t})$, and $\zeta_{k,t}$ denotes the selected samples on the $k$-th worker at the $t$-th iteration for the outer-level function. After obtaining the momentum $\mathbf{u}_{k, t}$, the $k$-th worker uses it to update the model parameter $\mathbf{x}_{k, t}$, which is shown below:
\begin{equation} \label{update_x}
	\begin{aligned}
		& \tilde{\mathbf{x}}_{k, t+1} = \sum_{j: w_{kj}> 0}w_{kj}\mathbf{x}_{j,t} - \gamma_x \mathbf{u}_{k, t} \ ,   \mathbf{x}_{k, t+1}= \mathbf{x}_{ k,t} + \eta(\tilde{\mathbf{x}}_{k, t+1}-\mathbf{x}_{ k,t} ) \ ,
	\end{aligned}
\end{equation}
where $\gamma_x>0$, $\eta \in (0, 1)$, and $w_{kj}$ denotes the edge weight between the $k$-th worker and the $j$-th worker. Here, $\sum_{j: w_{kj}> 0}w_{kj}\mathbf{x}_{j,t}$ denotes the communication with the neighoring workers  to get their model parameters. Then,  the $k$-th worker  updates the communicated model parameter with the momentum $\mathbf{u}_{k, t}$. The second step in Eq.~(\ref{update_x}) is the linear combination between $\mathbf{x}_{k,t}$ and the intermediate model parameter $\tilde{\mathbf{x}}_{k, t+1}$. 

Similar to the update of $\mathbf{x}_{k,t}$, each worker uses a similar strategy to update the model parameter  $\mathbf{y}_{k,t}$ in the maximization subproblem. It is worth noting that the maximization subproblem is not a compositional problem. Thus, we use the standard  stochastic gradient ascent with momentum algorithm to update $\mathbf{y}_{k,t}$, which can be found in Steps 4-5 of Algorithm~\ref{alg_dscgdamgp}.

\subsection{Theoretical Analysis}
We provide the convergence rate of our D-SCGDAM-GP algorithm below and the proof can be found in Appendix. 
\begin{theorem} \label{theorem_1}
	Under Assumptions~\ref{assumption_smooth}-\ref{assumption_graph}, by setting  $\beta_x>0$, $\beta_y>0$, $\alpha>0$,  $\eta< \min\{\frac{1}{\alpha}, \frac{1}{\beta_x}, \frac{1}{\beta_y}, \frac{1}{2\gamma_x L_{\Phi}}, 1\}$,  $\gamma_x\leq \min\{\frac{\gamma_y\mu^2} {20 C_g^2L_f^2}, \frac{\mu}{8L_f\sqrt{\gamma_{x, 1}}}, \frac{1-\lambda}{4\sqrt{\gamma_{x, 2}}} \}$, $\gamma_y\leq \min\{\frac{1}{6L_f}, \frac{1-\lambda}{3L_f\sqrt{\hat{C}_5 + 1 + 32/\beta_x^2 + 400/\beta_y^2}}, \frac{ 9\mu}{8L_f^2(8/\beta_x^2+100/\beta_y^2)} \}$, where $\gamma_{x, 1}=(8/\beta_x^2 )C_f^2L_g^2+  (104/\alpha + 315/\alpha^2+8/\beta_x^2+ 100/\beta_y^2) C_g^4L_f^2$, $\gamma_{x, 2}=\hat{C}_4+ C_f^2L_g^2 +    1264 C_g^4L_f^2 + (32/\beta_x^2 )C_f^2L_g^2+  4(104/\alpha + 315/\alpha^2+8/\beta_x^2+ 100/\beta_y^2) C_g^4L_f^2$, $\hat{C}_4=(2566 + 64/\beta_x^2 + 800/\beta_y^2+832/\alpha + 2520/\alpha^2) L_f^2C_g^4+ (5+64/\beta_x^2)  C_f^2L_g^2$, $\hat{C}_5= 55+ 64/\beta_x^2 + 800/\beta_y^2 $, 
	D-SCGDAM-GP in Algorithm~\ref{alg_dscgdamgp} has the  convergence rate:
	\vspace{-5pt}
	\begin{equation} \label{eq_convergence_1}
		\small
		\begin{aligned}
			&    \frac{1}{T}\sum_{t=0}^{T-1}(\mathbf{E}[ \|\nabla \Phi(\bar{\mathbf{x}}_{t})\|^2] + C_g^2L_f^2 \mathbf{E}[\|\mathbf{y}^*(\bar{\mathbf{x}}_{t}) -\bar{\mathbf{y}}_{t}\|^2])  \leq  \frac{2( \Phi({\mathbf{x}}_{0})- \Phi({\mathbf{x}}_{*}))}{\eta\gamma_xT} \\
			& +\frac{12 C_g^2L_f^2}{\eta T\gamma_y\mu}\mathbf{E}[\|\bar{\mathbf{y}}_{0}   - \mathbf{y}^{*}({\mathbf{x}}_0)\| ^2]   + O\Big(\frac{  L_f^2}{\eta T\beta_x\mu^2  }\Big) +  O\Big( \frac{L_f^2}{\eta T\beta_y\mu^2 }\Big)
			+ O\Big(\frac{L_f^2  }{\eta T\alpha\mu^2} \Big) +  O\Big(\frac{\eta\beta_x }{K}  \Big)
			\\
			&    + O\Big( \frac{\eta \beta_y L_f^2}{\mu^2K }\Big)   +   O\Big(\frac{\alpha\eta L_f^2 }{\mu^2K}\Big )  + O\Big(\frac{ \beta_x\eta  L_f^2}{\mu^2  } \Big) + O\Big(\frac{ \beta_y\eta L_f^2}{  \mu^2}\Big)  +  O\Big(\frac{   \alpha\eta L_f^2 }{ \mu^2}  \Big) +   O\Big( \frac{ \alpha^2\eta^2L_f^2}{\mu^2  }\Big) \  . \\
		\end{aligned}
	\end{equation}

\end{theorem}

\begin{remark}
	From Theorem~\ref{theorem_1}, it can be seen that  the dependence of $\gamma_x$ and $\gamma_y$ over the spectral gap and condition number is  in the order of  $O(\frac{1-\lambda}{\kappa^3})$ and $O(\frac{1-\lambda}{\kappa})$, respectively. Additionally,  it is easy to know $\beta_x=O(1)$, $\beta_y=O(1)$, and $\alpha=O(1)$. 
\end{remark}

\begin{corollary} \label{corollary_1}
	Under Assumptions~\ref{assumption_smooth}-\ref{assumption_graph},  by setting $\beta_x=O(1)$, $\beta_y=O(1)$,  $\alpha=O(1)$,  $\gamma_x=O(\frac{1-\lambda}{\kappa^3})$,  $\gamma_y = O(\frac{1-\lambda}{\kappa})$,   $\eta=O(\frac{\epsilon^2}{\kappa^2})$,  and $T=\frac{\kappa^5}{(1-\lambda)\epsilon^4}$,  D-SCGDAM-GP in Algorithm~\ref{alg_dscgdamgp} can achieve the $\epsilon$-accuracy solution: $\frac{1}{T}\sum_{t=0}^{T-1}(\mathbf{E}[\|\nabla \Phi(\bar{\mathbf{x}}_{t})\|^2] + C_g^2L_f^2\mathbf{E}[\|\mathbf{y}^*(\bar{\mathbf{x}}_{t}) -\bar{\mathbf{y}}_{t}\|^2] ) \leq \epsilon^2$.
\end{corollary}

\begin{remark}
	The dependence of $\eta$ on the condition number is caused by the term with the factor $L_f^2/\mu^2$, e.g., $O({ \beta_x\eta  L_f^2}/{\mu^2  } )$. 
\end{remark}

\begin{remark}
	To achieve the $\epsilon$-accuracy solution, the iteration  (communication) complexity  of D-SCGDAM-GP  is $O(\frac{\kappa^5}{(1-\lambda)\epsilon^4})$, and the   sample complexity on each worker is $O(\frac{\kappa^5}{(1-\lambda)\epsilon^4})$, which indicates D-SCGDAM-GP cannot achieve the linear speedup regarding the number of workers.

\end{remark}

\subsection{Discussion} \label{sec_communication_challenge}
From the convergence upper bound in Eq.~(\ref{eq_convergence_1}), it is easy to find that there are three terms that are in the order of  $O(\eta)$ and are not scaled by the number of workers $K$. As a result, the learning rate $\eta$ cannot be set to $O(\frac{K\epsilon^2}{\kappa^2})$ to achieve the linear speedup. In fact, those three terms are introduced by the consensus error about the momentum $\mathbf{E}[\| \mathbf{u}_{k,t} -  \bar{\mathbf{u}}_{t}\|^2]$, $\mathbf{E}[\| \mathbf{v}_{k,t} -  \bar{\mathbf{v}}_{t}\|^2]$, and the inner-level function $\mathbf{E}[\| \mathbf{h}_{k,t} -  \bar{\mathbf{h}}_{t}\|^2]$ (See Lemmas~5, 6, 7). Taking the inner-level function as an example, via the recursive expansion in terms of Lemma~5, we have $\mathbf{E}[\| \mathbf{h}_{k,t} -  \bar{\mathbf{h}}_{t}\|^2]\propto O(\eta)$. Such a large consensus error makes D-SCGDAM-GP fail to achieve the linear speedup.  
In other words, D-SCGDAM-GP is not sufficient to address \textit{the  challenge about the communication across workers caused by the compositional structure in the loss function}. 
Therefore, to achieve this issue, \textit{a feasible way is to reduce the consensus error for those three terms, e.g., having a higher-order dependence on the learning rate. }


\begin{algorithm}[h]
	\caption{D-SCGDAM-GT}
	\label{alg_dscgdamgt}
	\begin{algorithmic}[1]
		\REQUIRE $\mathbf{x}_0$, $\mathbf{y}_0$, $\eta\in (0, 1)$, $\gamma_x>0$, $\gamma_y>0$, $\beta_x>0$, $\beta_y>0$, $\alpha>0$,  $\alpha\eta \in (0, 1)$, $\beta_x\eta \in (0, 1)$, $\beta_y\eta \in (0, 1)$. 
		$\mathbf{p}_{k,-1}=\mathbf{0}$ ,  $\mathbf{q}_{k,-1}=\mathbf{0}$ ,$\mathbf{u}_{k,-1}=\mathbf{0}$ ,  $\mathbf{v}_{k,-1}=\mathbf{0}$
		
		\FOR{$t=0,\cdots, T-1$, each worker $k$} 
		
		\IF {$t==0$}
		\STATE  
		$\mathbf{x}_{k,0}=\mathbf{x}_0$ ,  $\mathbf{y}_{k,0}=\mathbf{y}_0$ ,
		$\mathbf{r}_{k, 0}=\mathbf{h}_{k, 0}= g_k(\mathbf{x}_{k, 0};  \xi_{k, 0})$ , \\
		$\mathbf{u}_{k, 0}=\nabla g_k(\mathbf{x}_{k, 0};  \xi_{k, 0})^T\nabla_{g}  f_k(\mathbf{h}_{k,0}, \mathbf{y}_{k,0}; \zeta_{k,0})$, 
		$\mathbf{v}_{k,0}=\nabla_{y}  f_k(\mathbf{h}_{k,0}, \mathbf{y}_{k,0}; \zeta_{k,0})$,
		\ENDIF
		\STATE 
		$\mathbf{p}_{k, t} = \sum_{j: w_{kj}> 0}w_{kj}\mathbf{p}_{j,t-1} + \mathbf{u}_{k, t}- \mathbf{u}_{k, t-1}$ , \\
		$\tilde{\mathbf{x}}_{k, t+1} = \sum_{j: w_{kj}> 0}w_{kj}\mathbf{x}_{j,t} - \gamma_x \mathbf{p}_{k, t}$ , \ 
		$\mathbf{x}_{k, t+1}= \mathbf{x}_{ k,t} + \eta(\tilde{\mathbf{x}}_{k, t+1}-\mathbf{x}_{ k,t} )$ , 
		\STATE 
		$\mathbf{q}_{k, t} = \sum_{j: w_{kj}> 0}w_{kj}\mathbf{q}_{j,t-1} + \mathbf{v}_{k, t}- \mathbf{v}_{k, t-1}$ ,\\
		$\tilde{\mathbf{y}}_{k, t+1}  = \sum_{j: w_{kj}> 0}w_{kj}\mathbf{y}_{j,t}+ \gamma_y \mathbf{q}_{k, t}$ , \ 
		$\mathbf{y}_{k, t+1}= \mathbf{y}_{ k,t} + \eta(\tilde{\mathbf{y}}_{k, t+1} -\mathbf{y}_{ k,t} ) $  ,
		\STATE 
		$\mathbf{h}_{k, t+1} = (1- \alpha\eta) \mathbf{h}_{k, t} +  \alpha\eta g_k(\mathbf{x}_{k,t+1};  \xi_{k, t+1})$ , \\
		\colorbox{pink}{$\mathbf{r}_{k, t+1} = \sum_{j: w_{ij}> 0}w_{kj}\mathbf{r}_{j,t} + \mathbf{h}_{k, t+1}- \mathbf{h}_{k, t}$}, \\
		$\mathbf{u}_{k, t+1} = (1-\beta_x\eta)\mathbf{u}_{k, t} + \beta_x\eta \nabla g_k(\mathbf{x}_{k, t+1};  \xi_{k, t+1})^T\nabla_{g}  f_k(\colorbox{pink}{$\mathbf{r}_{k,t+1}$}, \mathbf{y}_{k,t+1}; \zeta_{k,t+1})$,\\
		$\mathbf{v}_{k, t+1} = (1-\beta_y\eta)\mathbf{v}_{k, t} + \beta_y\eta\nabla_{y}  f_k(\colorbox{pink}{$\mathbf{r}_{k,t+1}$}, \mathbf{y}_{k,t+1}; \zeta_{k,t+1})$,
		
		\ENDFOR
	\end{algorithmic}
\end{algorithm}

\vspace{-10pt}
\section{Gradient-Tracking-based  Algorithm}
\vspace{-10pt}
In this section, we develop a novel decentralized algorithms for Eq.~(\ref{loss}) to address the communication challenge as discussed in Section~\ref{sec_communication_challenge}.


\subsection{Algorithmic Design: Reducing Consensus Errors}
As discussed in Section~\ref{sec_communication_challenge}, the large consensus error regarding the momentum and inner-level function makes Algorithm~\ref{alg_dscgdamgp} fail to achieve  linear speedup. To address this issue, 
in Algorithm~\ref{alg_dscgdamgt}, we develop a new algorithm, i.e.,  the gradient-tracking-based decentralized  stochastic compositional gradient descent ascent with momentum (D-SCGDAM-GT) algorithm. Generally speaking, D-SCGDAM-GT employs the gradient-tracking communication strategy to communicate the momentum and inner-level function to reduce the consensus error. It is worth noting that this is \textit{the first work communicating the inner-level function} to address the  challenge caused by the compositional structure. We believe this novel algorithmic design can be applied to other decentralized compositional optimization problems, such as  the compositional minimization problem. 

In detail, Algorithm~\ref{alg_dscgdamgt} uses the same way as Algorithm~\ref{alg_dscgdamgp} to compute the momentum $\mathbf{u}_{k, t}$ for the stochastic compositional gradient $\nabla g_k(\mathbf{x}_{k, t};  \xi_{k, t})^T\nabla_{g}  f_k(\mathbf{r}_{k,t}, \mathbf{y}_{k,t}; \zeta_{k,t})$ and $\mathbf{v}_{k, t}$ for the stochastic gradient $\nabla_{y}  f_k(\mathbf{r}_{k,t}, \mathbf{y}_{k,t}; \zeta_{k,t})$, which is demonstrated in Step 5 of Algorithm~\ref{alg_dscgdamgt}.  However, the stochastic gradient $\nabla_{g}  f_k(\mathbf{r}_{k,t}, \mathbf{y}_{k,t}; \zeta_{k,t})$ and $\nabla_{y}  f_k(\mathbf{r}_{k,t}, \mathbf{y}_{k,t}; \zeta_{k,t})$ are computed based on the tracked estimator $\mathbf{r}_{k,t}$ regarding the inner-level function, rather than the original inner-level function estimator $\mathbf{h}_{k,t}$.  Specifically, $\mathbf{r}_{k,t}$ is updated with the gradient-tracking communication strategy:
\begin{equation}
	\mathbf{r}_{k, t+1} = \sum_{j: w_{kj}> 0}w_{kj}\mathbf{r}_{j,t} + \mathbf{h}_{k, t+1}- \mathbf{h}_{k, t} \ . 
\end{equation} 
Due to this additional communication step, the consensus error regarding $\mathbf{E}[\| \mathbf{r}_{k,t} -  \bar{\mathbf{r}}_{t}\|^2]$ is supposed to be smaller than the consensus error $\mathbf{E}[\| \mathbf{h}_{k,t} -  \bar{\mathbf{h}}_{t}\|^2]$ in Algorithm~\ref{alg_dscgdamgp}, which can facilitate the linear speedup \footnote{Here, we use $\mathbf{r}_{k,t}$ to compute stochastic gradients rather than $\mathbf{h}_{k,t}$ so that we should consider the consensus error about $\mathbf{r}_{k,t}$ in Algorithm~\ref{alg_dscgdamgt}, rather than $\mathbf{h}_{k,t}$. }. In fact, this is confirmed by Lemma~19,
where we can have $\mathbf{E}[\| \mathbf{r}_{k,t} - \bar{\mathbf{r}}_{t}\|^2] \propto O(\eta^2)$ via the recursive expansion. When $\eta=O(\epsilon^2)$, this consensus error is much smaller than   $\mathbf{E}[\| \mathbf{h}_{k,t} -  \bar{\mathbf{h}}_{t}\|^2]\propto O(\eta)$ obtained from Lemma~5.

After obtaining the momentum, each worker $k$ applies the gradient tracking communication strategy to the momentum $\mathbf{u}_{k, t}$ and $\mathbf{v}_{k, t}$ to reduce the consensus error regarding the momentum, which is shown in Step 5 of Algorithm~\ref{alg_dscgdamgt}. Then, the tracked momentum $\mathbf{p}_{k, t}$ and $\mathbf{q}_{k, t}$ are used to  update the model parameter $\mathbf{x}_{k, t}$ and $\mathbf{y}_{k, t}$, respectively.

\vspace{-5pt}
\subsection{Theoretical Analysis}
\vspace{-5pt}

\begin{theorem} \label{theorem_2}
	Under Assumptions~\ref{assumption_smooth}-\ref{assumption_graph}, by setting  $\beta_x>0$, $\beta_y>0$, $\alpha>0$,  $\eta< \min\{\frac{1}{\alpha}, \frac{1}{\beta_x}, \frac{1}{\beta_y}, \frac{1}{2\gamma_x L_{\Phi}}, 1\}$, $\gamma_x\leq \min\{ \frac{\mu^2\gamma_y}{20 C_g^2L_f^2} , \frac{\mu (1-\lambda)}{8L_f\sqrt{\gamma_{x, 1}}}\}$, $\gamma_y \leq \min\{\frac{1}{6L_f},  \frac{9\mu}{8L_f^2(8/\beta_x^2 + 100/\beta_y^2)}  \}$, where $\gamma_{x, 1}=8(C_g^4L_f^2+ C_f^2L_g^2 )/\beta_x^2 +  100C_g^4L_f^2/\beta_y^2  +  312C_g^4L_f^2/\alpha + 4056C_g^4L_f^2$, 
	Algorithm~\ref{alg_dscgdamgt} has the  convergence rate:
	\begin{equation}
		\small
		\begin{aligned}
			&  \frac{1}{T}\sum_{t=0}^{T-1}(\mathbf{E}[ \|\nabla \Phi(\bar{\mathbf{x}}_{t})\|^2] +  C_g^2L_f^2\mathbf{E}[\|\mathbf{y}^*(\bar{\mathbf{x}}_{t}) -\bar{\mathbf{y}}_{t}\|^2] ) \leq \frac{2(\Phi({\mathbf{x}}_{0})- \Phi({\mathbf{x}}_{*}))}{\eta\gamma_x T} \\
			&  + \frac{12C_g^2L_f^2}{\eta T\gamma_y\mu}\mathbf{E}[\|\bar{\mathbf{y}}_{0}   - \mathbf{y}^{*}(\bar{\mathbf{x}}_0)\| ^2]     + O\Big( \frac{1}{\eta T\beta_x} \Big)  +  O\Big(\frac{L_f^2}{\eta T\beta_y\mu^2} \Big) +  O\Big( \frac{L_f^2}{\eta T\mu^2} \Big)+  O\Big(\frac{\beta_x\eta}{K} \Big) \\
			&  +  O\Big(\frac{ \alpha\eta L_f^2}{ T\mu^2(1-\lambda)^2}\Big)   +O\Big(\frac{ L_f^2}{T(1-\lambda) \mu^2} \Big) +   O\Big(\frac{\beta_y\eta L_f^2}{K\mu^2}\Big) +O\Big(\frac{ \gamma_x^2   \eta^2 L_f^2 }{(1-\lambda)^4\mu^2}\Big) + O\Big(\frac{ \gamma_x^2   \beta_x^2\eta^2  L_f^2}{\beta_y^2\mu^2(1-\lambda)^4}   \Big)  \\
			&    +  O\Big( \frac{\eta\alpha^2 L_f^2}{K\mu^2} \Big) + O\Big(\frac{ \gamma_x^2   \beta_x^2\eta^4L_f^2}{\mu^2(1-\lambda)^6} \Big)+ O\Big(\frac{\gamma_y^2\beta_y^2\eta^2 L_f^2}{\mu^2(1-\lambda)^4} \Big)+   O\Big(\frac{\gamma_y^2\beta_y^2\eta^2 L_f^2}{\beta_x^2\mu^2(1-\lambda)^4}\Big)  + O\Big(\frac{\gamma_y^2\eta^2 L_f^2}{\mu^2(1-\lambda)^4}\Big) \\
			&   +  O\Big(\frac{\alpha^2\eta^2 L_f^2}{(1-\lambda)^2 \mu^2}\Big) +   O\Big(\frac{ \alpha^3\eta^3  L_f^2}{\mu^2(1-\lambda)^2}\Big)   +  O\Big(\frac{ \gamma_x^2   \beta_x^2\eta^2L_f^2}{(1-\lambda)^5\mu^2}\Big ) + O\Big( \frac{ \gamma_x^2   \beta_x^2\eta^2 L_f^2 }{\alpha\mu^2(1-\lambda)^4}  \Big)\ .  \\
		\end{aligned}
	\end{equation}

\end{theorem}

\begin{remark}
	From Theorem~\ref{theorem_2}, it can be seen that  the dependence of  $\gamma_x=O(\frac{1-\lambda}{\kappa^3})$,  $\gamma_y=O(\frac{1}{\kappa})$, $\beta_x=O(1)$, $\beta_y=O(1)$,  and $\alpha=O(1)$. Compared with Theorem~\ref{theorem_1}, $\gamma_y$ does not depend on the spectral gap $1-\lambda$. 
\end{remark}

\begin{corollary} \label{corollary_2}
	Under Assumptions~\ref{assumption_smooth}-\ref{assumption_graph},  by setting $\beta_x=O(1)$, $\beta_y=O(1)$,   $\alpha=O(1)$,  $\gamma_x=O(\frac{1-\lambda}{\kappa^3})$,  $\gamma_y=O(\frac{1}{\kappa})$, $\eta=O(\frac{K\epsilon^2}{\kappa^2})$, $T=O(\frac{\kappa^5}{(1-\lambda)\epsilon^4K})$,  D-SCGDAM-GT in Algorithm~\ref{alg_dscgdamgt} can achieve the $\epsilon$-accuracy solution.

\end{corollary}

\begin{remark}
	To achieve the $\epsilon$-accuracy solution, the iteration (communication) complexity is $O(\frac{\kappa^5}{(1-\lambda)\epsilon^4K})$, which  indicates the linear speedup regarding the number of workers $K$ and is better than Algorithm~\ref{alg_dscgdamgp}. Additionally,  the sample complexity on each worker is $O(\frac{\kappa^5}{(1-\lambda)\epsilon^4K})$, which  also indicates  the linear speedup.
\end{remark}


\section{Experiment}

\subsection{Experimental Setup}

\textbf{Datasets.} In our experiments, we will use our methods to optimize Eq.~(\ref{compositional_auc}).   We use three benchmark datasets: catvsdog \footnote{\url{https://www.kaggle.com/c/dogs-vs-cats}}, stl10 \cite{coates2011analysis}, fashionmnist \cite{xiao2017/online}. Specifically, we  split the dataset into two groups where the first half of classes compose the first group and the other half of classes compose the second group. Then, for one of these two groups, we randomly eliminate some examples such that these two groups form an imbalanced binary classification dataset. In our experiments, the ratio between the number of positive samples and all samples is set to $0.1$. In addition, the ratio between the training and testing set is $9:1$.  
Then, the dataset is randomly distributed to all engaged workers. Each worker uses the obtained dataset to compute stochastic (compositional) gradients to update local model parameters. 

\begin{figure*}[!h]
	\centering 
	\subfigure[catvsdog]{
		\includegraphics[scale=0.3085]{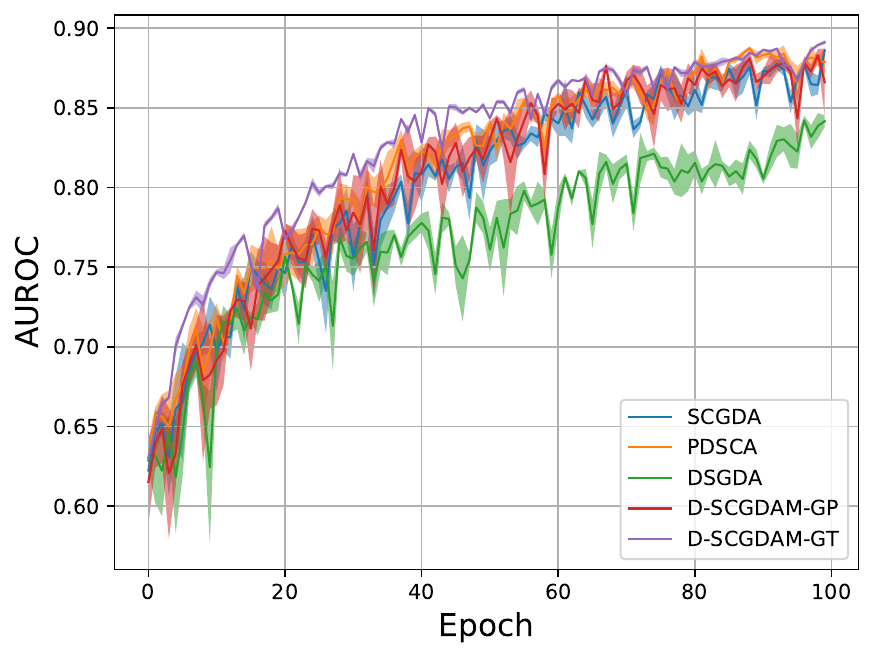}
	}
	\hspace{-12pt}
	\subfigure[stl10]{
		\includegraphics[scale=0.3085]{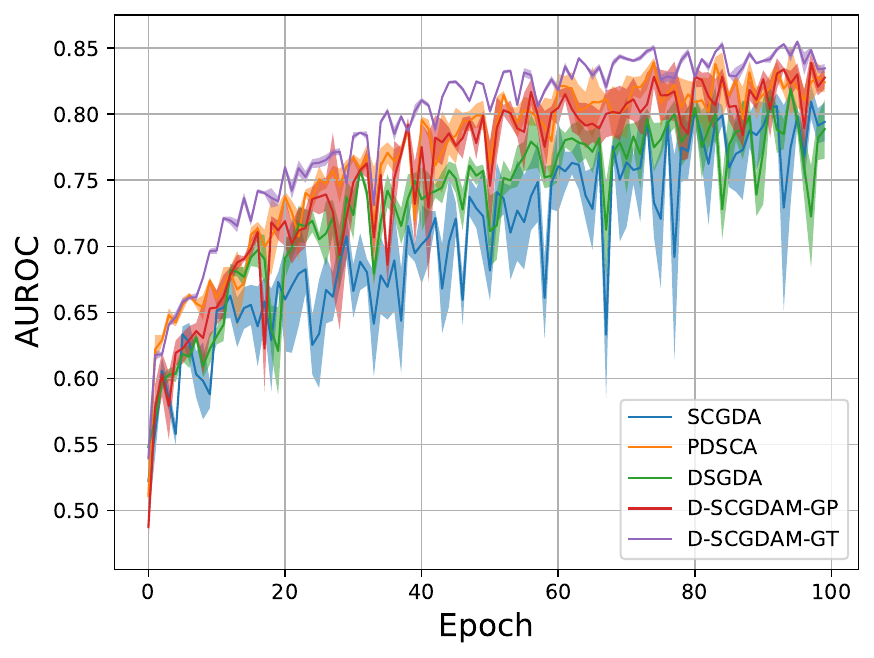}
	}
	\hspace{-12pt}
	\subfigure[fashionmnist]{
		\includegraphics[scale=0.3085]{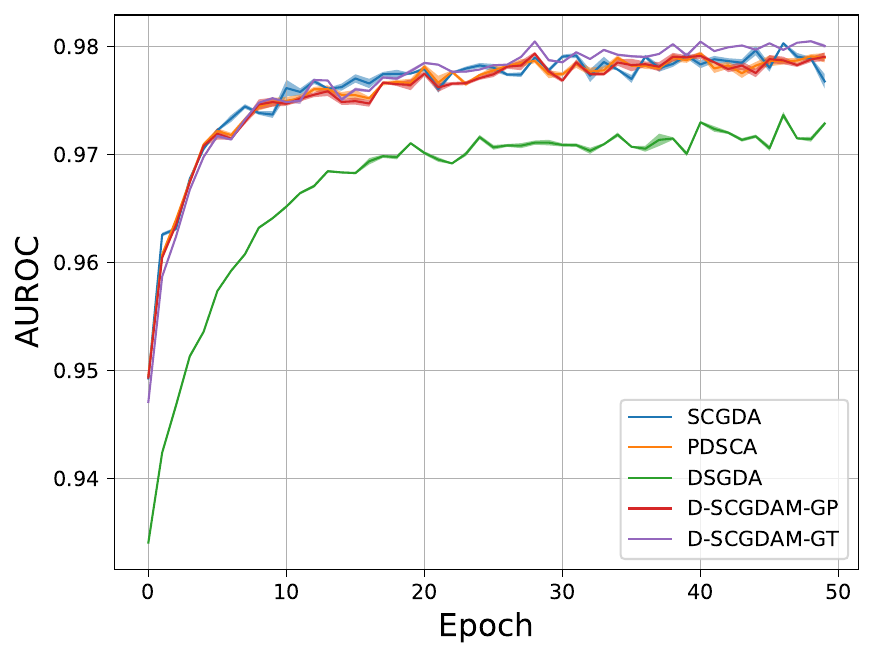}
	}
	\caption{The comparison of the test AUROC score between our algorithms and baseline algorithms. }
	\label{testauc}
\end{figure*}


\textbf{Settings.} The classifier for imaging datasets is ResNet20 where the last layer is revised to perform binary classification, while the classifier for tabular datasets is an MLP, which has one hidden layer with 16 neurons.  Since  our method is the first parallel method for the compositional minimax problem, there are no counterpart methods. Thus, in our experiment, we compare our methods with the single-machine method: SCGDA \cite{gao2021convergence} and PDSCA \cite{yuan2021compositional}, as well as a non-compositional decentralized stochastic gradient descent ascent (DSGDA) method \cite{tsaknakis2020decentralized}.  Note that, to make a fair comparison with SCGDA and PDSCA, we parallelize these two baseline methods  based on a fully connected communication graph, i.e., $1-\lambda=1$. The reason is that batch normalization used in  ResNet20 for imaging datasets is sensitive to the batch size. Thus, to make a fair comparison, we should enforce all methods are trained with the same batch size on each worker.  
The hyperparameters of our methods are set as: $\gamma_x=\gamma_y=0.99$, $\beta_x=\beta_y=9.9$, $\alpha=9.0$, $\eta=0.1$, $\rho=0.1$. As for baseline methods, we use the same learning rate with our methods. Finally, all methods are implemented with PyTorch and MPI.

\subsection{Result and Analysis}
In Figure~\ref{testauc}, we plot the test AUROC score of our methods and baseline methods. In this experiment, we use four workers and the communication graph is a ring graph. The batch size on each worker is 32 for catvsdog, fashionmnist, 8 for stl10,  512 for credit fraud, 128 for w8a and ijcnn1. Here, the test AUROC score is computed upon the averaged model $\bar{\mathbf{x}}$. 
From Figure~\ref{testauc}, it can be seen that our two algorithms can converge to almost the same test AUROC score with the single-machine momentum-based algorithm PDSCA, which confirms the correctness of our algorithms. Additionally, our algorithms can achieve a  better AUROC score than DSGDA,  confirming the effectiveness of  compositional gradient. 

\begin{wrapfigure}[12]{R}{2.4in}
	\vspace{-15pt}
	\centering
	\includegraphics[scale=0.4]{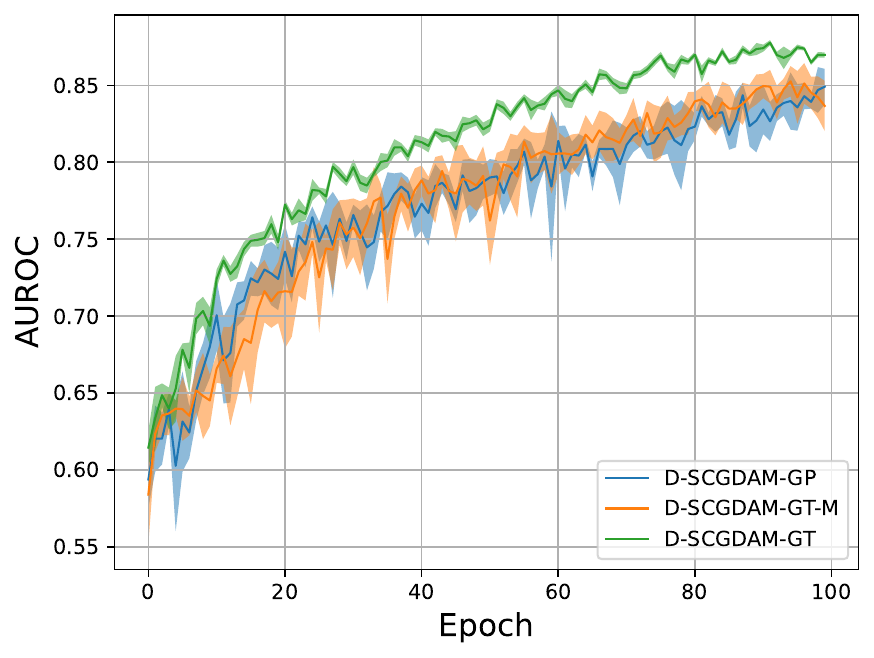}
	\vspace{-0.05in}
	\caption{The test AUROC score  for catvsdog when using the mini-batch size 16.}
	\label{fig_batchsize}
\end{wrapfigure} 

To demonstrate the capability of D-SCGDAM-GT in controlling the consensus error, we use a small mini-batch size, i.e., 16, so that the stochastic gradient and stochastic inner-level function value on each worker are more noisy than the case with a large batch size, i.e., 32. The other settings do not change.  In Figure~\ref{fig_batchsize}, we plot the test AUROC score for CATvsDOG dataset, where D-SCGDAM-GT-M means we apply gradient tracking to the momentum but not the inner-level function. It can be seen that D-SCGDAM-GT-M performs slightly better than D-SCGDAM-GP, but much worse than D-SCGDAM-GT. This confirms the necessity of communicating the inner-level function for improving the convergence performance.

\section{Conclusion}
In this paper, we have developed two novel decentralized stochastic compositional gradient descent ascent methods, demonstrating how to achieve linear speedup for stochastic compositional minimax optimization problems. The superior experimental performance confirms the correctness and effectiveness of our methods.  All these theoretical and empirical results corroborate the novelty and contribution of our work. 

\bibliographystyle{abbrv}
\bibliography{egbib}


\end{document}